\documentclass[a4paper]{article}

\usepackage{arxiv}

\usepackage[utf8]{inputenc} 
\usepackage{hyperref}       
\usepackage{url}            
\usepackage{booktabs}       
\usepackage{amsfonts}       
\usepackage{graphicx}       
\usepackage{subcaption}
\usepackage{algorithm}
\usepackage{algorithmicx}
\usepackage{algpseudocode}
\usepackage{amsmath}
\usepackage{multirow}
\usepackage{a4}
\usepackage{enumitem}
\usepackage{amssymb}
\usepackage{amsthm}
\usepackage{mathtools}
\usepackage{float}
\usepackage{longtable}

\graphicspath{ {./images/} }

\title{EXAdam: The Power of Adaptive Cross-Moments}

\author{
 Ahmed M. Adly\\
 Independent Researcher \& Software Engineer \\
 Egypt \\
}

\begin{document}
\maketitle

\begin{abstract}
    This paper introduces EXAdam (\textbf{EX}tended \textbf{Adam}), a novel optimization algorithm that builds upon the widely-used Adam~\cite{kingma2014adam} optimizer. EXAdam incorporates two key enhancements: (1) new debiasing terms for improved moment estimation and (2) a gradient-based acceleration mechanism for increased responsiveness to the current loss landscape. These innovations work synergistically to address limitations of the original Adam algorithm, potentially offering improved convergence properties, enhanced ability to escape saddle points, and potentially greater robustness to hyperparameter choices, though this requires further investigation. We provide a theoretical analysis of EXAdam's components and their interactions, highlighting the algorithm's potential advantages in navigating complex optimization landscapes. Empirical evaluations demonstrate EXAdam's superiority over Adam, achieving 38.46\% faster convergence and yielding improvements of 1.96\%, 2.17\%, and 1.17\% in training, validation, and testing accuracies, respectively, when applied to a CNN trained on the CIFAR-10 dataset~\cite{krizhevsky2014cifar}. While these results are promising, further empirical validation across diverse tasks is essential to fully gauge EXAdam's efficacy. Nevertheless, EXAdam represents a significant advancement in adaptive optimization techniques, with promising implications for a wide range of machine learning applications. This work aims to contribute to the ongoing development of more efficient, adaptive, and universally applicable optimization methods in the field of machine learning and artificial intelligence.
\end{abstract}

\section{Introduction}

Optimization is a fundamental problem in machine learning, where the goal is to minimize a loss function that measures the difference between the model's predictions and the true labels. Stochastic gradient descent (SGD) and its variants are widely used optimization algorithms in deep learning, due to their simplicity, computational efficiency, and ability to handle large datasets.

However, SGD has its limitations, particularly when dealing with noisy or high-variance gradients, which can lead to slow and unstable convergence. To address these challenges, adaptive gradient methods have been proposed, but their effectiveness is still a topic of debate. For instance, Wilson et al.~\cite{wilson2018marginalvalueadaptivegradient} found that adaptive gradient methods tend to generalize less effectively than SGD with momentum across a range of deep learning tasks, including image classification, character-level language modeling, and constituency parsing.

Different hypotheses about the origins of this worse generalization have been investigated, such as the presence of sharp local minima~\cite{keskar2017largebatchtrainingdeeplearning, dinh2017sharpminimageneralizedeep} and inherent problems of adaptive gradient methods~\cite{wilson2018marginalvalueadaptivegradient}. These findings highlight the need for a deeper understanding of the underlying mechanisms driving the performance of adaptive gradient methods.

In seeking such understanding, researchers have identified momentum as a crucial element in many iterative optimization algorithms. Momentum has been consistently shown to accelerate and improve convergence, as demonstrated by Nemirovskii and Nesterov~\cite{nemirovskii1986OptimalMO}, and frequently yields solutions with enhanced generalization capabilities, as found by Sutskever et al.\cite{sutskever2013OnTI}. Through the accumulation of gradient vectors across successive optimization steps, momentum facilitates the overcoming of minor local fluctuations in the loss landscape, potentially escaping shallow local minima and accelerating progress in plateau regions, as discussed by Qian\cite{qian1999momentum}, Ruder~\cite{ruder2016overview}, and Goh~\cite{goh2017WhyMR}.

In recent years, adaptive gradient algorithms such as Adam\cite{kingma2014adam}, RMSprop\cite{tieleman2012rmsprop}, and AMSGrad~\cite{reddi2019convergence} have gained popularity due to their ability to adapt to the geometry of the loss function and stabilize the optimization process. Among these, Adam is one of the most widely used optimizers, known for its simplicity, computational efficiency, and good performance in practice. However, Adam has some limitations, such as its bias towards the initial iterations and its sensitivity to the choice of hyperparameters.

To address these limitations, we propose a new optimization algorithm, called EXAdam, which enhances the traditional Adam optimizer by incorporating novel debiasing terms and some additional components. In my experiments, we show that EXAdam outperforms traditional Adam, achieving 38.46\% faster convergence and yielding 1.96\% higher training accuracy and 2.17\% higher validation accuracy on a CNN trained on the CIFAR-10 dataset~\cite{krizhevsky2014cifar} compared to Adam. Similarly, it achieved 28.89\% faster convergence, 1.25\% higher training accuracy, and 0.62\% higher validation accuracy compared to AdamW~\cite{loshchilov2019decoupledweightdecayregularization}.

The primary motivation of this paper is to enhance Adam's performance with EXAdam, making it a more competitive choice compared to traditional Adam and other state-of-the-art optimizers, even in scenarios where Adam previously struggled. Our goal is to provide practitioners with a reliable and versatile optimization algorithm that can adapt to a wide range of problems, eliminating the need to switch between different optimizers and hyperparameters for specific datasets or tasks. By doing so, we hope to simplify the optimization process and reduce the burden of hyperparameter tuning, ultimately leading to more efficient and effective deep-learning model development.

\section{Methods}

\subsection{New Debiasing Terms}

The Adam optimizer, introduced in 2014, is a popular stochastic gradient descent algorithm that adapts the learning rate for each parameter based on the magnitude of the gradient \cite{kingma2014adam}. The original Adam optimizer uses debiasing terms $\hat{m}$ and $\hat{v}$ to correct for the bias in the estimates of the first and second moments of the gradient, respectively. However, these terms have limitations that can affect the convergence and stability of the optimization process \cite{reddi2019convergence}. Specifically, the original Adam optimizer's debiasing terms treat the first and second moments independently, failing to account for their mutual influence on parameter updates \cite{loshchilov2019decoupledweightdecayregularization}. This independence assumption can lead to suboptimal scaling of updates, particularly in regions of high curvature where the interaction between gradient magnitude and variance is crucial \cite{luo2019AdaptiveGM}. Additionally, the fixed nature of the debiasing terms doesn't adapt to the local geometry of the loss landscape, potentially resulting in either overly aggressive or conservative parameter updates.

In the pursuit of improving the Adam optimizer, we introduce a novel approach to debiasing, which we term $\tilde{m}$ and $\tilde{v}$. These new terms aim to rectify the inherent bias in the traditional Adam update rules, thereby enhancing the overall performance and stability of the optimizer.

The traditional Adam optimizer relies on the debiased estimates $\hat{m}$ and $\hat{v}$, which are computed as $\hat{m} = \frac{m}{1 - \beta_1^t}$ and $\hat{v} = \frac{v}{1 - \beta_2^t}$, respectively. While these terms effectively mitigate the bias introduced by the exponential moving averages, they still suffer from limitations. Specifically, $\hat{m}$ and $\hat{v}$ do not fully account for the interplay between the first and second moments, leading to suboptimal convergence.

To address these limitations, we propose the novel debiasing terms $\tilde{m}$ and $\tilde{v}$, defined as in Equation \ref{eq:debiasing_terms}.

\begin{equation}
    \label{eq:debiasing_terms}
    \tilde{m} = \frac{m}{1 - \beta_1^t} \left(1 + \frac{v}{v + \epsilon}\cdot \beta_2^t\right) \quad \text{and} \quad \tilde{v} = \frac{v}{1 - \beta_2^t} \left(1 + \frac{m^2}{m^2 + \epsilon}\cdot \beta_1^t \right)
\end{equation}

where $m$ and $v$ are the first- and second-moment estimates of the gradient, respectively, $\beta_1$ and $\beta_2$ are the exponential decay rates for the moment estimates, $\epsilon$ is a small constant to prevent division by zero, and $t$ is the current iteration number.

These terms are designed to capture the intricate relationships between the first and second moments, thereby providing a more accurate and nuanced representation of the gradient statistics.

The $\tilde{m}$ term can be seen as a refinement of the traditional $\hat{m}$ estimate. By incorporating the second moment $v$ and the learning rate $\beta_2^t$, $\tilde{m}$ better accounts for the variance of the gradient, leading to more informed updates. The additional term $\left(1 + \frac{v}{v + \epsilon}\cdot \beta_2^t\right)$ serves as a correction factor, which adaptively scales the debiasing process based on the relative magnitude of the variance.

Similarly, the $\tilde{v}$ term builds upon the traditional $\hat{v}$ estimate, incorporating the first moment $m$ and the scaling factor $\beta_1^t$. This allows $\tilde{v}$ to more effectively capture the covariance between the gradient and its squared value, resulting in a more accurate estimate of the variance.

A closer look at the $\tilde{m}$ and $\tilde{v}$ terms reveals several advantages. One key benefit is that they reduce the bias in the optimization process more gradually, thanks to the adaptive correction factors that temper the scaling factors $\beta_1^t$ and $\beta_2^t$. This leads to a more stable and robust optimization process.

The inclusion of $\beta_1^t$ and $\beta_2^t$ in $\tilde{v}$ and $\tilde{m}$ respectively introduces a time-dependent factor that evolves as training progresses. This temporal component allows the bias correction to adapt dynamically throughout the optimization process, potentially offering improved long-term stability and convergence properties. As $t$ approaches infinity, the $\beta_1^t$ and $\beta_2^t$ approach 0. Consequently, the new terms asymptotically converge to the original Adam bias correction terms. This property ensures that the long-term behavior of the algorithm remains well-defined and consistent with the original formulation.

The cross-moment interaction and temporal dynamics are particularly impactful in the early stages of training when $t$ is small. During this critical phase, the new terms provide a more nuanced correction that may lead to improved initial convergence and stability.

The debiasing terms also have a profound impact on the optimization process. When the gradients are noisy, indicated by a large second moment estimate $v$, the correction factor is closer to 1, which means the bias correction is more aggressive. This makes sense, as we want to correct for the bias more strongly when the gradients are unreliable. On the other hand, when the gradients are stable, indicated by a small second moment estimate $v$, the correction factor is closer to 0, which means the bias correction is less aggressive. This also makes sense, as we don't want to over-correct when the gradients are consistent.

Similarly, when the gradients have a strong direction, indicated by a large first-moment estimate $m$, the correction factor is closer to 1, which means the bias correction is more aggressive. This is intuitive, as we want to correct the bias more strongly when the gradients have a clear direction. Conversely, when the gradients are weak, indicated by a small first moment estimate $m$, the correction factor is closer to 0, which means the bias correction is less aggressive. This also makes sense, as we don't want to over-correct when the gradients are uncertain.

One notable aspect of the new debiasing terms in EXAdam is the intentional asymmetry in treating the first moment estimate $m$ and the second moment estimate $v$. Specifically, $v$ appears unsquared in the correction term for $\tilde{m}$, while $m$ is squared in the correction term for $\tilde{v}$. This design choice is grounded in the statistical properties of these moments.

The first-moment estimate $m$ represents the mean of the gradients and is on the same scale as the gradients themselves. In contrast, the second-moment estimate $v$ represents the uncentered variance of the gradients, which is naturally on a squared scale. To maintain consistency with these inherent scales, we use $v$ and $m^2$ in their respective correction terms. This ensures scale consistency, which is essential for robust optimization.

The use of $v$ in the correction term for $\tilde{m}$ provides a more immediate and sensitive response to the current gradient variability. This variance-based scaling factor allows the first-moment estimate to be more reactive to changes in the gradient landscape. In contrast, using $m^2$ in the correction term for $\tilde{v}$ provides a more stable correction. The squaring operation ensures that the correction is always positive and gives more weight to larger gradient values, which can help stabilize the second-moment estimate, especially in scenarios with sparse or highly variable gradients.

The asymmetric design also allows each moment estimate to benefit from complementary information. By using different forms of the moments ($v$ and $m^2$), we incorporate distinct aspects of the optimization trajectory into each correction term. This can lead to more robust and adaptive behavior. Furthermore, the use of $m^2$ in the correction term for $\tilde{v}$ creates a normalization effect that is bounded between 0 and 1. This provides a well-behaved scaling factor that can smoothly adjust the second moment estimate without risk of extreme values.

The asymmetric design in EXAdam reflects a nuanced approach to moment estimation and bias correction. By tailoring the correction terms to the statistical nature of each moment estimate, EXAdam aims to balance the need for responsive updates with the requirement for stable, long-term learning. This approach potentially allows EXAdam to achieve a more sophisticated adaptation to the underlying optimization landscape.

The interplay between the first and second moments is also more effectively captured, allowing the optimizer to better adapt to the underlying gradient landscape. This is particularly important in scenarios where the gradient variance is high, as the $\tilde{m}$ and $\tilde{v}$ terms provide a more nuanced understanding of the gradient statistics. Coupling the first and second moment estimates allows the optimizer to adapt to the complex dynamics of the optimization process.

Let me give you an example to illustrate how these novel debiasing terms work in practice. Imagine training a neural network to navigate a mountain landscape, where the network needs to find the lowest valley (global minimum). The traditional Adam optimizer with $\hat{m}$ and $\hat{v}$ is like a hiker who makes decisions based on two separate pieces of information: the slope direction ($\hat{m}$ - first moment) and the terrain roughness ($\hat{v}$ - second moment). EXAdam, with $\tilde{m}$ and $\tilde{v}$, is like a more sophisticated hiker who makes smarter decisions by considering how these factors interact.

When the hiker encounters:
\begin{itemize}
    \item \textbf{Rough, steep terrain} (high variance, strong gradient): $\tilde{m}$ increases the correction because $\frac{v}{v + \epsilon}$ is larger. This makes EXAdam more cautious but decisive when it has a clear direction in challenging terrain
    \item \textbf{Smooth, gentle slope} (low variance, weak gradient): Both correction terms become smaller as $\frac{v}{v + \epsilon}$ and $\frac{m^2}{m^2 + \epsilon}$ decrease. The optimizer can take more confident steps as the terrain is more predictable
\end{itemize}

The asymmetric design (using $v$ vs $m^2$) is particularly clever. Using the raw $v$ for $\tilde{m}$ is like the hiker being immediately responsive to terrain changes, while using $m^2$ for $\tilde{v}$ is like maintaining a more stable, long-term assessment of the path.

What makes this especially powerful is the temporal component ($\beta_1^t$ and $\beta_2^t$). Early in training (small $t$), these corrections have more influence, like a hiker being extra cautious when starting on unfamiliar terrain. As training progresses (large $t$), the terms gradually approach the original Adam behavior, similar to a hiker becoming more confident after gaining experience with the landscape.


\subsection{Gradient-based Acceleration Mechanism}

In this section, we introduce a novel acceleration mechanism for the Adam optimizer, which leverages gradient information to enhance the convergence rate of the algorithm. This innovation is rooted in the observation that the gradient itself contains valuable information about the optimization landscape, which can be harnessed to accelerate the optimization process. To achieve this, we propose a novel component, termed the gradient-based acceleration mechanism, denoted by $\tilde{g}$. This acceleration mechanism is designed to synergistically interact with the debiased momentum $\tilde{m}$, enabling the optimizer to more effectively harness the gradient information in a controlled manner and accelerate the convergence process. The gradient-based acceleration mechanism $\tilde{g}$ is defined as in Equation \ref{eq:gradient_accelerator}.

\begin{equation}
    \label{eq:gradient_accelerator}
    \tilde{g} = \frac{g}{1 - \beta_1^t} \left(1 + \frac{v}{v + \epsilon} \cdot \beta_2^t \right)
\end{equation}

where $g$ is the gradient, $\beta_1$ and $\beta_2$ are the exponential decay rates for the moment estimates, $\epsilon$ is a small constant to prevent division by zero, and $t$ is the current iteration number.

The gradient-based acceleration mechanism $\tilde{g}$ can be interpreted as a measure of the gradient's "urgency" or "importance". When the gradient is large, this means the surface is steep, and the optimizer is far from the optimal solution, thus, it should prioritize updating the parameters in the direction of the gradient. In this case, $\tilde{g}$ increases, indicating that the optimizer should focus on refining the parameters based on the momentum $\tilde{m}$. On the other hand, when the gradient is small, this means the surface is flat, and the optimizer should explore the parameter space more broadly. In this case, $\tilde{g}$ decreases, allowing the optimizer to focus on refining the parameters based on the momentum $\tilde{m}$.

When $\tilde{g}$ increases, it indicates that the optimizer should prioritize updating the parameters in the direction of the gradient. On the other hand, when the gradient is small, $\tilde{g}$ decreases, allowing the optimizer to focus on refining the parameters based on the momentum $\tilde{m}$.

The term $(1 + \frac{v}{v + \epsilon} \cdot \beta_2^t)$ provides an adaptive scaling to the gradient. This scaling factor is always greater than or equal to 1, with its magnitude determined by the second moment estimate $v$ and the current timestep $t$.

The incorporation of $v$ in the scaling term creates an interaction between the direct gradient and the second moment estimate. In regions of high curvature (large $v$), the gradient term receives additional emphasis, potentially allowing for more aggressive updates in these areas. Conversely, in regions of low curvature (small $v$), the gradient term is downweighted, enabling the optimizer to rely more on the momentum term for guidance.

The gradient-based acceleration mechanism $\tilde{g}$ is incorporated into the update rule as follows:

\begin{equation}
    \theta \leftarrow \theta - \frac{\alpha \cdot \left(\tilde{m} + \tilde{g}\right)}{\sqrt{\tilde{v}} + \epsilon}  
\end{equation}

where $\theta$ is the parameter vector, $\alpha$ is the learning rate, $\tilde{m}$ is the bias-corrected first-moment estimate, $\tilde{v}$ is the bias-corrected second-moment estimate, $\tilde{g}$ is the bias-corrected gradient, $\epsilon$ is a small constant to prevent division by zero, and $t$ is the current iteration number.

The combination of $\tilde{m}$ and $\tilde{g}$ in the update rule creates a highly adaptive learning mechanism. It balances the smoothed, historical information captured by $\tilde{m}$ with the immediate, potentially more volatile information in $\tilde{g}$. In scenarios where the loss landscape changes rapidly, the immediate responsiveness provided by $\tilde{g}$ could lead to faster convergence compared to methods that rely solely on smoothed estimates.

The direct incorporation of the current gradient, especially with its adaptive scaling, could potentially enhance the ability to escape saddle points or shallow local minima by providing more immediate directional information, although specific experiments targeting saddle point traversal were not performed in this study. This could be particularly beneficial in deep learning scenarios with complex loss landscapes.

The relative influence of $\tilde{m}$ and $\tilde{g}$ changes over time and is based on the local geometry of the loss surface. This dynamic trade-off could potentially offer the benefits of both momentum-based methods and more reactive gradient-based approaches.

A closer examination of the gradient-based acceleration mechanism $\tilde{g}$ reveals several desirable properties. Firstly, $\tilde{g}$ exhibits a self-correcting behavior, as the gradient information is adaptively scaled based on the variance $v$. This ensures that the optimizer remains robust to outliers and noisy gradients. Secondly, the incorporation of $\tilde{g}$ into the update rule enables the optimizer to more effectively adapt to changing gradient landscapes.

To illustrate how this gradient-based acceleration mechanism works in practice, imagine we're hiking down a mountain to reach a valley (the optimal solution), but we're doing this hike in foggy conditions. The traditional Adam optimizer is like having a compass (momentum $\tilde{m}$) and a topographical map (second moment estimate $\tilde{v}$). The modification of the gradient-based acceleration mechanism adds something like a real-time slope sensor ($\tilde{g}$) that actively measures how steep our current position is.

Let's consider three scenarios:

\begin{enumerate}
    \item \textbf{Steep Slope (Large Gradient):}
        When we're on a steep section of the mountain, the traditional Adam relies heavily on past movements (momentum), while EXAdam detects the steepness through $\tilde{g}$ and increases the influence of the current gradient. The term $\frac{v}{v + \epsilon} \cdot \beta_2^t$ becomes larger. This is like saying, "Hey, this is really steep, let's pay more attention to where we're currently stepping rather than just following our planned path."
    \item \textbf{Gentle Slope (Small Gradient):}
         When we're on a relatively flat section, the $\tilde{g}$ term becomes smaller and the optimizer relies more on the momentum term $\tilde{m}$. This is like saying, "The terrain is pretty flat, let's stick more to our general direction from the compass."
    \item \textbf{Mixed Terrain (Varying Curvature):}
         The really clever part comes in areas with varying steepness. If we hit a sudden, steep drop (high curvature), the $v$ term in the equation increases. This automatically scales up the gradient term through $(1 + \frac{v}{v + \epsilon} \cdot \beta_2^t)$. It's like having an adaptive hiking strategy that seamlessly switches between careful stepping on steep sections and momentum-based movement on flatter ground.
\end{enumerate}

Let me express this mathematically with a simple scenario:
$$
\text{At iteration } t = 100, \beta_2 = 0.999:
$$

For a steep region where $v = 0.1$:
$$
\tilde{g} = \frac{g}{1 - \beta_1^{100}} (1 + \frac{0.1}{0.1 + 10^{-8}} \cdot 0.999^{100}) \approx 2g
$$

For a flat region where $v = 0.001$:
$$
\tilde{g} = \frac{g}{1 - \beta_1^{100}} (1 + \frac{0.001}{0.001 + 10^{-8}} \cdot 0.999^{100}) \approx 1.1g
$$

This shows how this mechanism automatically provides stronger gradient acceleration in steeper regions while maintaining more conservative updates in flatter areas. This adaptive behavior could be particularly valuable in training deep neural networks, where the loss landscape can vary dramatically across different layers and during different phases of training.

The beauty of this modification is that it maintains the core benefits of Adam (adaptive learning rates and momentum) while adding this extra layer of responsiveness to the current landscape, potentially helping to navigate difficult optimization terrains more effectively.

\subsection{EXAdam Algorithm}

Let's put everything together and see the complete EXAdam algorithm, which seamlessly integrates all the enhancements discussed earlier. This unified approach combines the novel debiasing terms $\tilde{m}$ and $\tilde{v}$, and the gradient-based accelerator $\tilde{g}$. Algorithm \ref{alg:EXAdam} provides a comprehensive view of EXAdam, showcasing how these innovations work together to create a more adaptive and potentially powerful optimization method. This algorithmic representation encapsulates the essence of my contributions, offering a clear roadmap for implementing and further studying EXAdam's behavior in various optimization scenarios.

\begin{algorithm}[h]
\caption{EXAdam, an enhanced variant of the Adam optimizer}
\label{alg:EXAdam}
\begin{algorithmic}[1]
\State Initialize parameters $\theta$, first moment estimate $m = 0$, second moment estimate $v = 0$
\State Initialize hyperparameters $\alpha$, $\beta_1$, $\beta_2$, $\epsilon$
\For{iteration $t = 1, 2, \ldots, T$}
\State Compute gradient $g_t = \nabla_{\theta} L(\theta)$
\State Update first moment estimate: $m \leftarrow \beta_1 m + (1 - \beta_1) g_t$
\State Update second moment estimate: $v \leftarrow \beta_2 v + (1 - \beta_2) g_t^2$
\State Compute bias-corrected first moment estimate: $\tilde{m} = \frac{m}{1 - \beta_1^t} \left(1 + \frac{v}{v + \epsilon}\cdot \beta_2^t\right)$
\State Compute bias-corrected second moment estimate: $\tilde{v} = \frac{v}{1 - \beta_2^t} \left(1 + \frac{m^2}{m^2 + \epsilon}\cdot \beta_1^t \right)$
\State Compute gradient-based accelerator: $\tilde{g} = \frac{g_t}{1 - \beta_1^t} \left(1 + \frac{v}{v + \epsilon} \left(\beta_2^t\right)\right)$
\State Update parameters: $\theta \leftarrow \theta - \alpha \cdot\frac{\left(\tilde{m} + \tilde{g}\right)}{\sqrt{\tilde{v}}+ \epsilon}$
\EndFor
\end{algorithmic}
\end{algorithm}

\section{Experiments}

We empirically evaluated the effectiveness of EXAdam through a series of experiments on two diverse benchmark tasks. The first experiment involved training a convolutional neural network (CNN) on the CIFAR-10 dataset, a widely used benchmark for image classification tasks. This setup allowed us to assess EXAdam's performance in a relatively smooth gradient landscape. Then, we trained a MinGPT model on a dataset of Shakespeare works, a challenging task that requires the optimizer to navigate a complex and nuanced gradient landscape. This experiment enabled us to evaluate the optimizer's ability to handle long-range dependencies and adapt to changing gradient statistics. 

In each of these experiments, we compared the performance of EXAdam to state-of-the-art optimizers using identical hyperparameters and training protocols. By doing so, we aimed to assess the robustness and versatility of EXAdam across different model architectures, datasets, and tasks, and to demonstrate its ability to outperform existing state-of-the-art optimizers in a variety of settings. We implemented all the experiments using PyTorch~\cite{paszke2019pytorch} to ensure the compatibility and reproducibility of the results. The experiment was conducted on Kaggle. The code for the experiment is available on GitHub\footnote{\href{https://github.com/AhmedMostafa16/EXAdam}{The repository of EXAdam on GitHub}}.

We note that using the same learning rate for all optimizers facilitates comparison but may not be optimal for each individual algorithm. A comprehensive analysis of EXAdam's sensitivity to its hyperparameters ($\alpha$,$\beta_{1}$,$\beta_{2}$) compared to other optimizers is a considerable direction for future work.

\subsection{Experiment: Image Classification}

\begin{figure}[htbp]
    \centering
    \begin{subfigure}[b]{0.49\linewidth}
      \includegraphics[width=\linewidth]{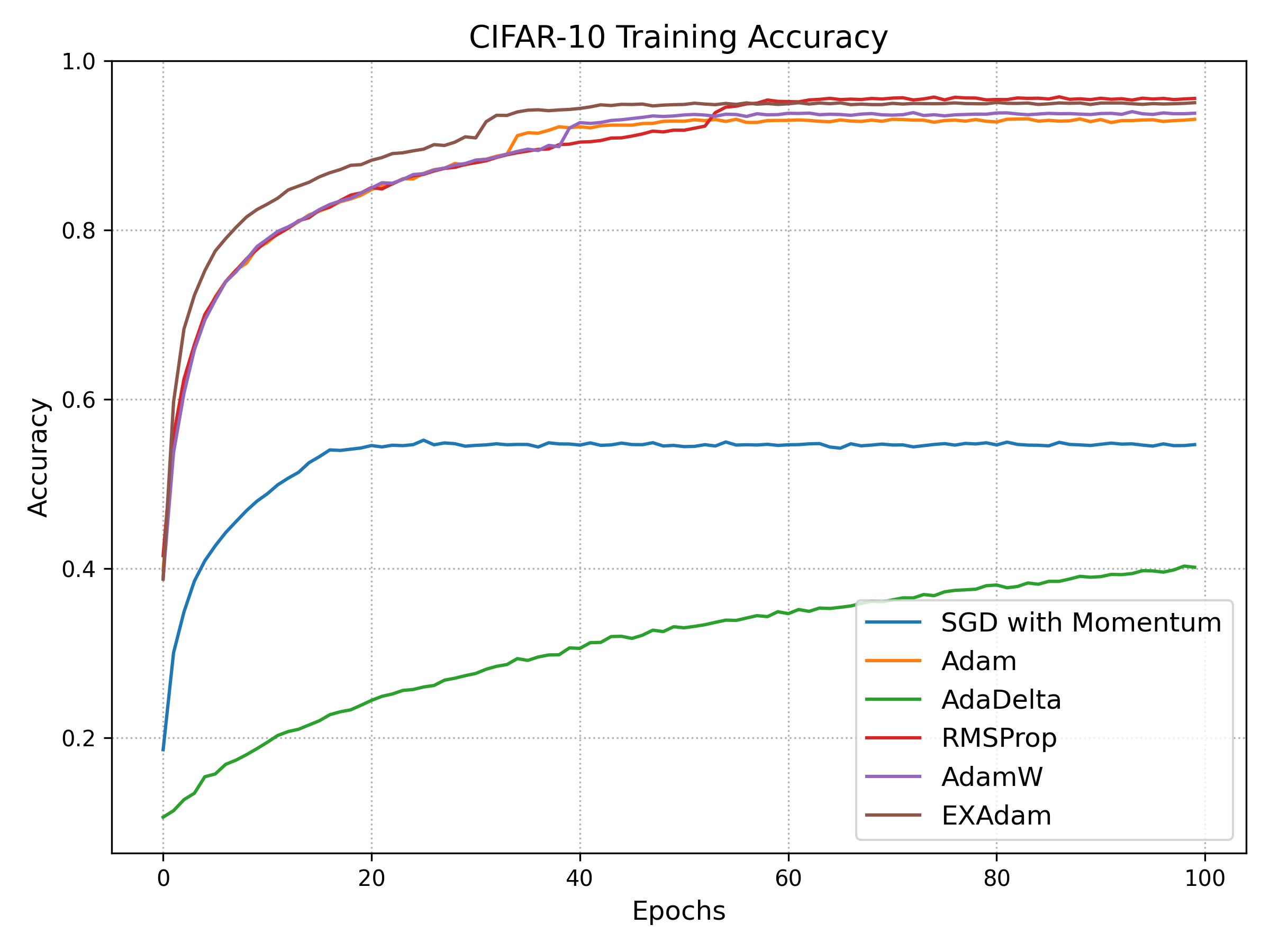}
      \caption{CIFAR-10 training accuracy}
    \end{subfigure}
    \begin{subfigure}[b]{0.49\linewidth}
      \includegraphics[width=\linewidth]{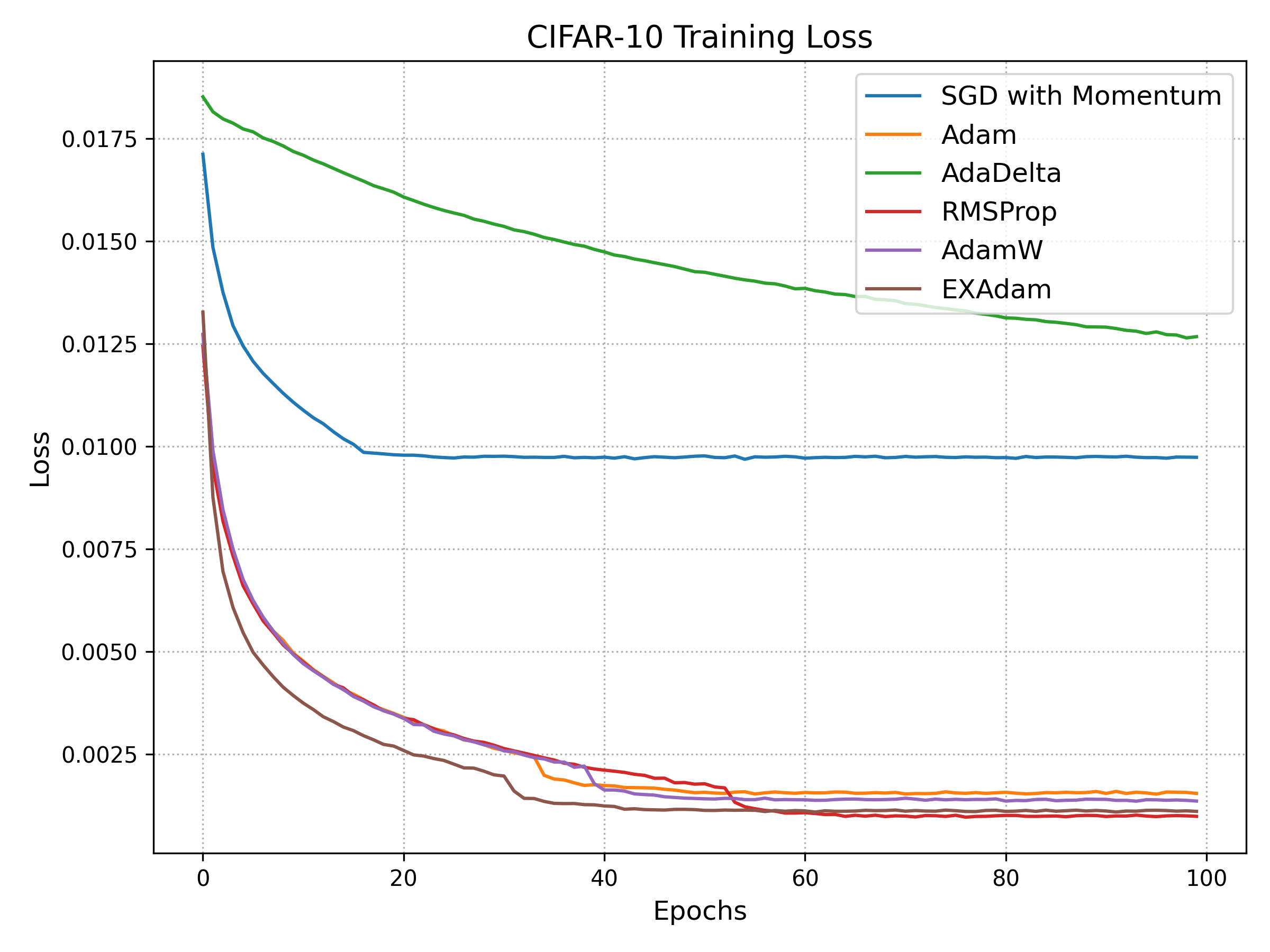}
      \caption{CIFAR-10 training loss}
    \end{subfigure}
    \caption{The training performance of EXAdam, Adam, AdamW, SGD with momentum, RMSProp, and AdaDelta on the CIFAR-10 dataset. The convexities in the training curves indicate that the \texttt{ReduceLROnPlateau} learning rate scheduler reduced the learning rate.}
    \label{fig:cifar10_training_performance}
\end{figure}

\begin{figure}[htbp]
    \centering
    \begin{subfigure}[b]{0.49\linewidth}
      \includegraphics[width=\linewidth]{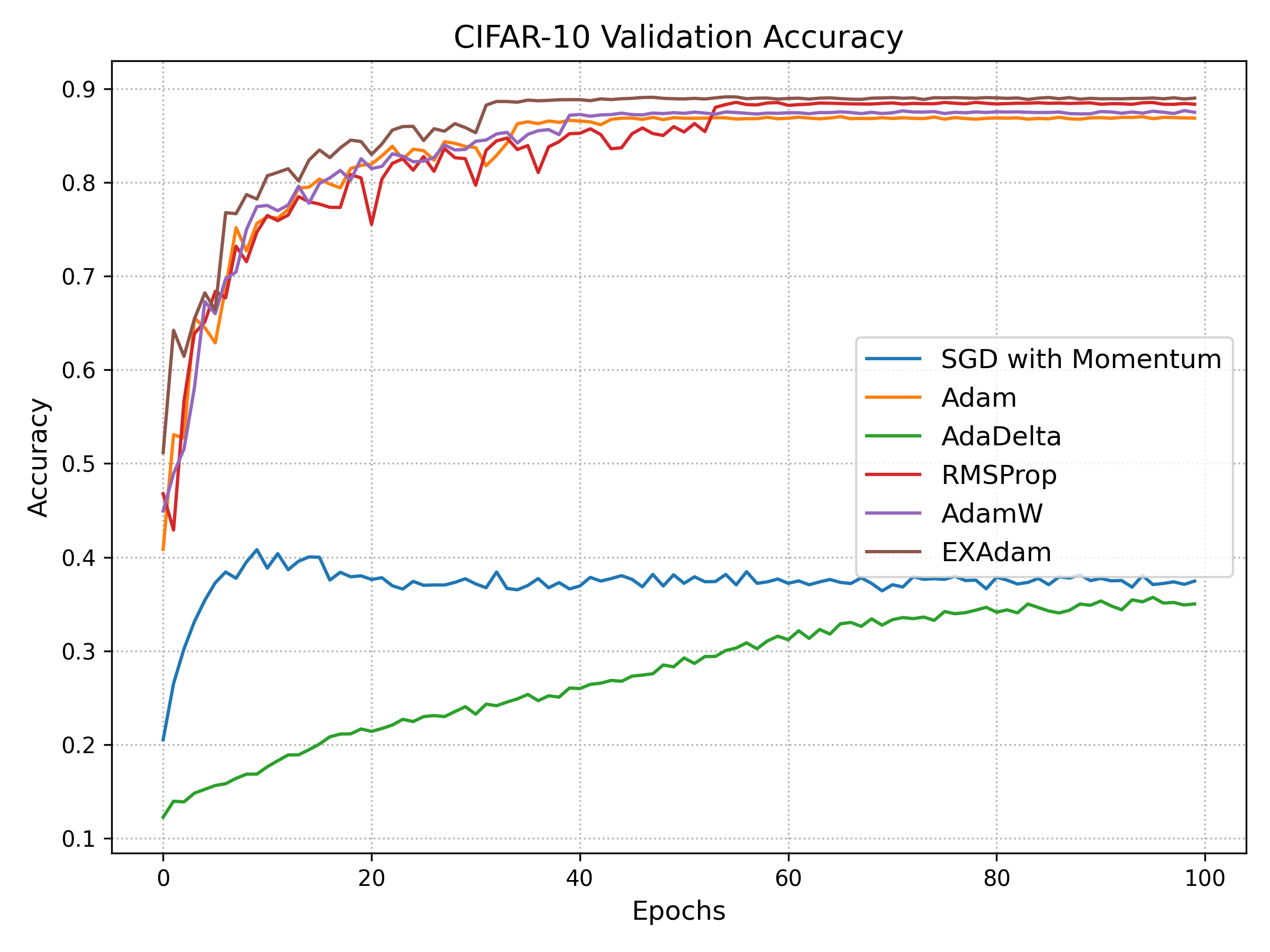}
      \caption{CIFAR-10 validation accuracy}
    \end{subfigure}
    \begin{subfigure}[b]{0.49\linewidth}
      \includegraphics[width=\linewidth]{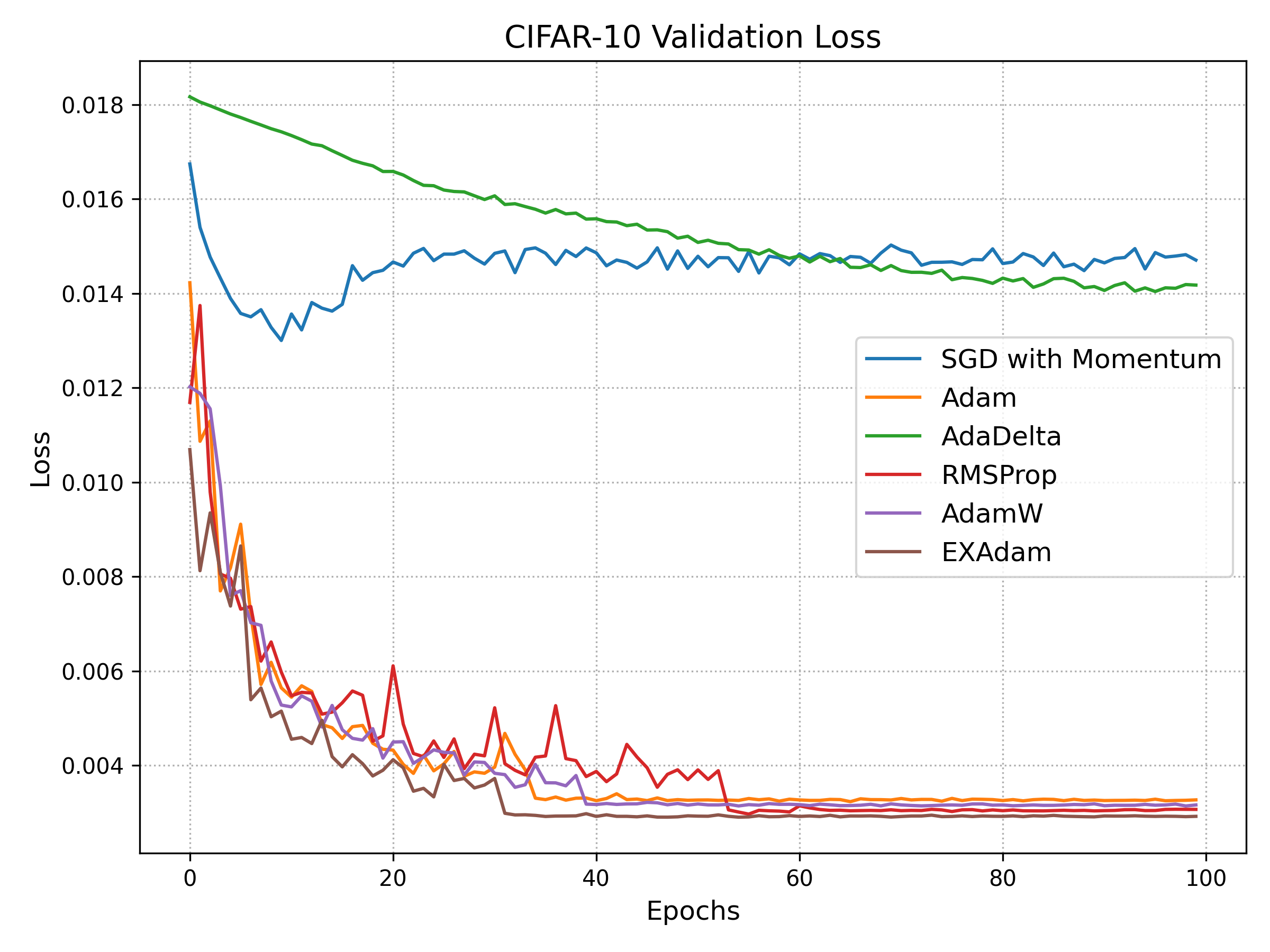}
      \caption{CIFAR-10 validation loss}
    \end{subfigure}
    \caption{The validation performance of EXAdam, Adam, AdamW, SGD with momentum, RMSProp, and AdaDelta on the CIFAR-10 dataset.}
    \label{fig:cifar10_validation_performance}
\end{figure}

We conducted a benchmark experiment involving image classification using the CIFAR-10 dataset. This experiment aimed to assess the convergence properties of EXAdam compared to traditional Adam, AdamW, AdaDelta, SGD with momentum, and RMSProp on a deep convolutional neural network (CNN) model. The goal was to determine whether EXAdam could achieve faster convergence and higher accuracy on the CIFAR-10 dataset, showcasing its effectiveness in practice. The results of the experiment are presented in Figure \ref{fig:cifar10_training_performance} and Figure \ref{fig:cifar10_validation_performance}. The values of the loss and accuracy at different epochs are shown in Table \ref{tab:loss_accuracy}.

The CNN model employed in this experiment is a deep neural network consisting of multiple convolutional and fully connected layers, making 3 million trainable parameters. The network takes as input a 3-channel 32x32 image and outputs a probability distribution over the 10 classes of the CIFAR-10 dataset.

The network architecture can be divided into three main blocks, each comprising a sequence of convolutional, batch normalization, and max pooling layers, followed by a residual connection. The first block consists of two convolutional layers with 64 filters, each with a kernel size of 3 and padding set to "same" to preserve spatial dimensions. The output of the second convolutional layer is fed into a residual block, which applies two convolutional layers with batch normalization and ReLU activation, followed by a shortcut connection that adds the input to the output. The output of the residual block is then max pooled with a stride of 2, reducing the spatial dimensions by half.

The second block is similar in structure to the first, with two convolutional layers with 128 filters, followed by a residual block and max pooling. The third block consists of two convolutional layers with 256 filters, followed by a residual block and max pooling.

After the third block, the output is flattened and fed into two fully connected layers, the first with 128 units and ReLU activation, and the second with the number of units equal to the number of classes in the dataset. Dropout regularization with a probability of 0.25 is applied after each max pooling layer and after the first fully connected layer.

The network was trained using the proposed EXAdam, Adam, AdamW, SGD with momentum, RMSProp, and AdaDelta~\cite{zeiler2012adadelta} to evaluate their performance. We compare the convergence behavior, training and validation loss, and accuracy of the models trained with EXAdam against those trained with other state-of-the-art optimizers. The number of training iterations was set to 100, and the learning rate was set to 0.0001 for all optimizers. The performance of the network was evaluated on the CIFAR-10 dataset, and the results are compared to those obtained using other optimizers.

Additionally, a learning rate scheduler was employed to dynamically adjust the learning rate during training. The scheduler, \texttt{ReduceLROnPlateau}, reduces the learning rate by a factor of 0.1 when the validation loss plateaus, with a patience of 5 epochs. The minimum learning rate was set to 0, and the scheduler was configured to operate in "min" mode, reducing the learning rate when the validation loss stops improving. The scheduler's verbose mode was enabled to provide detailed output during training. To ensure the reproducibility of the results, the random seed was set to 1234 before training the network. This seed was used to initialize the weights of the network and the random number generator, ensuring that the results are consistent across different runs.

To comprehensively evaluate EXAdam's performance, we analyzed its convergence behavior throughout training, tracking loss and accuracy at regular intervals. This detailed analysis allows us to compare EXAdam to other optimizers in terms of learning speed, stability, and generalization. We also examined gradient statistics and parameter updates to understand how EXAdam's novel components contribute to its performance, gaining valuable insights into its behavior and advantages in handling complex deep neural network optimization landscapes.

\begin{table}[htbp] 
\centering
\caption{CIFAR-10 CNN Training and Validation Loss and Accuracy at Different Epochs}
\label{tab:loss_accuracy}
\vspace{3pt} 

\resizebox{\textwidth}{!}{
\begin{tabular}{l c r r r r}
\toprule 
Optimizer Name & Epoch & Training Loss & Training Accuracy (\%) & Validation Loss & Validation Accuracy (\%) \\ 
\midrule 

\multirow{7}{*}{Adam}   & 1 & 0.01271 & 39.85 & 0.01423 & 40.84 \\
                        & 5 & 0.00673 & 69.58 & 0.00819 & 64.50 \\
                        & 10 & 0.00497 & 77.89 & 0.00564 & 75.62 \\
                        & 25 & 0.00307 & 86.07 & 0.00389 & 83.53 \\
                        & 50 & 0.00156 & 92.91 & 0.00326 & 86.90 \\
                        & 75 & 0.00158 & 92.77 & 0.00324 & 86.96 \\
                        & 100 & 0.00155 & 93.12 & 0.00327 & 86.85 \\
\midrule 
\multirow{7}{*}{AdamW}  & 1 & 0.01274 & 38.96 & 0.01202 & 44.93 \\
                        & 5 & 0.00676 & 69.39 & 0.00760 & 67.28 \\
                        & 10 & 0.00493 & 78.07 & 0.00528 & 77.41 \\
                        & 25 & 0.00299 & 86.59 & 0.00433 & 82.19 \\
                        & 50 & 0.00142 & 93.51 & 0.00316 & 87.43 \\
                        & 75 & 0.00139 & 93.66 & 0.00316 & 87.55 \\
                        & 100 & 0.00136 & 93.83 & 0.00316 & 87.47 \\
\midrule 
\multirow{7}{*}{AdaDelta}& 1 & 0.01852 & 10.64 & 0.01817 & 12.26 \\
                        & 5 & 0.01774 & 15.42 & 0.01780 & 15.23 \\
                        & 10 & 0.01719 & 18.74 & 0.01743 & 16.86 \\
                        & 25 & 0.01575 & 25.73 & 0.01628 & 22.46 \\
                        & 50 & 0.01426 & 33.15 & 0.01521 & 28.30 \\
                        & 75 & 0.01336 & 36.82 & 0.01449 & 33.26 \\
                        & 100 & 0.01268 & 40.17 & 0.01418 & 35.01 \\
\midrule 
\multirow{7}{*}{RMSProp}& 1 & 0.01244 & 41.55 & 0.01169 & 46.77 \\
                        & 5 & 0.00661 & 70.04 & 0.00797 & 65.11 \\
                        & 10 & 0.00494 & 77.73 & 0.00598 & 74.68 \\
                        & 25 & 0.00302 & 86.43 & 0.00452 & 81.30 \\
                        & 50 & 0.00177 & 91.83 & 0.00370 & 85.94 \\
                        & 75 & 0.00098 & 95.74 & 0.00306 & 88.38 \\
                        & 100 & 0.00099 & 95.58 & 0.00307 & 88.33 \\
\midrule 
\multirow{7}{*}{SGD with Momentum}& 1 & 0.01712 & 18.61 & 0.01675 & 20.52 \\
                        & 5 & 0.01245 & 40.93 & 0.01389 & 35.40 \\
                        & 10 & 0.01108 & 47.98 & 0.01300 & 40.80 \\
                        & 25 & 0.00973 & 54.66 & 0.01470 & 37.42 \\
                        & 50 & 0.00976 & 54.58 & 0.01453 & 38.12 \\
                        & 75 & 0.00974 & 54.69 & 0.01466 & 37.71 \\
                        & 100 & 0.00974 & 54.67 & 0.01471 & 37.45 \\
\midrule 
\multirow{7}{*}{\textbf{EXAdam}}  & 1 & 0.01327 & 38.77 & 0.01069 & 51.16 \\
                        & 5 & 0.00547 & 75.22 & 0.00738 & 68.20 \\
                        & 10 & 0.00393 & 82.44 & 0.00515 & 78.21 \\
                        & 25 & 0.00235 & 89.39 & 0.00334 & 85.97 \\
                        & 50 & 0.00115 & 94.84 & 0.00293 & 88.92 \\
                        & 75 & 0.00114 & 94.96 & 0.00292 & 89.03 \\
                        & 100 & 0.00111 & 95.08 & 0.00292 & 89.02 \\
\bottomrule 
\end{tabular}
} 
\end{table}

As shown in the results, EXAdam outperformed all the other tested state-of-the-art optimizers in terms of training and validation accuracy. EXAdam achieved higher accuracy on the CIFAR-10 dataset during training and validation, therefore, better testing accuracy can be expected. The testing accuracy of EXAdam, Adam, AdamW, SGD with momentum, RMSProp, and AdaDelta is shown in Table \ref{tab:testing_accuracy}.

\begin{table}[htbp]
  \centering
\caption{Testing Accuracy on CIFAR-10 Dataset}
\vspace{3pt}
\label{tab:testing_accuracy}
\begin{tabular}{c|c}
\hline
Optimizer Name & Testing Accuracy (\%) \\
\hline
Adam & 89.66 \\
AdamW & 89.71 \\
AdaDelta & 37.45 \\
RMSProp & 90.14 \\
SGD with Momentum & 43.13 \\
EXAdam & 90.83 \\
\hline
\end{tabular}
\end{table}

The superior performance of EXAdam, as presented in Table \ref{tab:testing_accuracy}, is further evidenced by its testing accuracy, which surpasses all other optimizers by a significant margin. EXAdam achieves a 90.83\% testing accuracy, outperforming RMSProp (90.14\%), Adam (89.66\%), and AdamW (89.71\%) by 1.96, 2.44, and 2.39 percentage points, respectively. Notably, AdaDelta and SGD with Momentum demonstrate significantly lower testing accuracy, with scores of 37.45\% and 43.13\%, respectively. 

This substantial increase in accuracy demonstrates EXAdam's enhanced generalization capability, likely resulting from its novel debiasing terms and adaptive learning rate mechanism. The improved testing accuracy suggests that EXAdam not only converges faster during training but also learns more robust and generalizable features from the dataset. This performance gain is particularly noteworthy given the challenging nature of the CIFAR-10 dataset and the competitive baseline set by Adam and AdamW, which are widely regarded as state-of-the-art optimizers.

Compared to the original Adam and AdamW optimizers, EXAdam introduces a small number of additional element-wise operations per step to compute the new debiasing and acceleration terms. This results in a ~2.5\% increase in computational overhead during training this CNN, which is negligible in most cases.

The results underscore EXAdam's potential as a powerful optimization algorithm for deep learning tasks, especially in scenarios where achieving high accuracy on complex datasets is crucial.

\subsection{Experiment: Text Generation}

\begin{figure}[htbp]
    \centering
    \includegraphics[width=\textwidth]{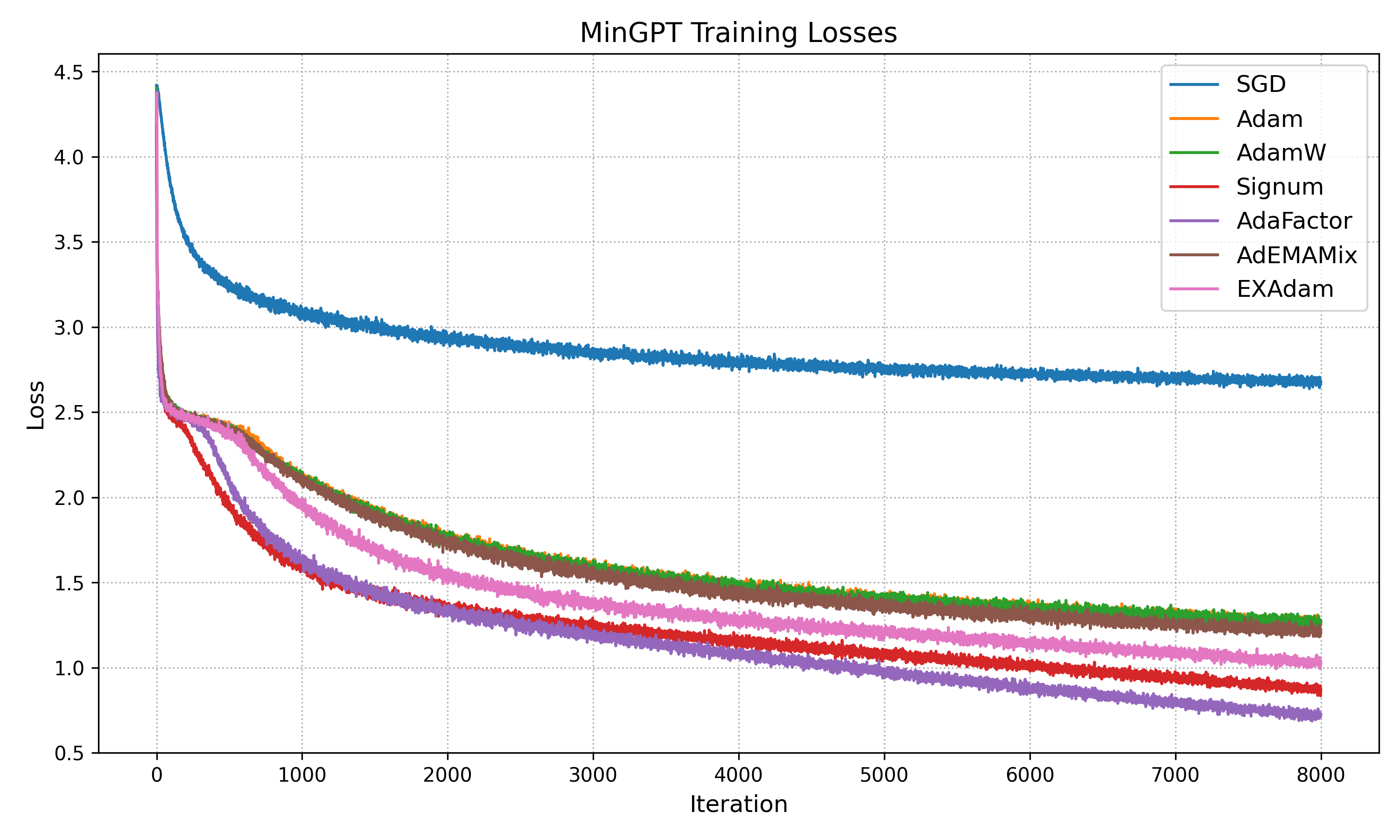}
    \caption{Training loss of MinGPT using EXAdam, Adam, AdamW, AdaFactor, SGD with Momentum, AdEMAMix, and Signum. The loss curves show the convergence behavior of the optimizers during training on the Shakespeare dataset.}
    \label{fig:mingpt_training_performance}
\end{figure}

We conducted another experiment to evaluate the performance of EXAdam in the context of text generation by training a 14.3-million parameter MinGPT model, a mini version of the GPT (Generative Pre-trained Transformer) model~\cite{radford2018improving}, on a dataset of Shakespeare works\footnote{\href{https://www.kaggle.com/datasets/kadonis/shakespeare6}{Shakespeare dataset on Kaggle}}. The goal of this experiment was to assess the optimizer's ability to handle long-range dependencies and complex loss landscapes, which are common challenges in natural language processing tasks. The results of the experiment are shown in Figure \ref{fig:mingpt_training_performance}.

This language model is based on the transformer architecture~\cite{vaswani2023attentionneed} and consists of an embedding layer, a series of transformer blocks, and a final linear layer. The embedding layer maps each input token to a dense vector, which is then fed into the transformer blocks. Each transformer block consists of self-attention and feed-forward neural network (FFNN) layers. The self-attention layer computes the weighted sum of the input tokens, while the FFNN layer transforms the output of the self-attention layer. The final linear layer outputs the logits for each token in the vocabulary.

We used the Shakespeare dataset, which contains a relatively good amount of text that could fit in the memory of the GPUs we had access to. We split the data into training and validation sets, with 90\% of the data used for training and 10\% for validation. We create a mapping from characters to indices and vice versa, and define encoding and decoding functions to convert between characters and indices.

In addition to EXAdam, we also experimented with several other optimizers to compare their performance on the MinGPT model. Specifically, we evaluated Adam, AdamW, AdaFactor~\cite{shazeer2018adafactor}, SGD with Momentum, AdEMAMix~\cite{pagliardini2024ademamix}, and Signum~\cite{bernstein2018signsgd}, each with their default hyperparameters. These optimizers were chosen because they have been widely used in deep learning and have shown promising results in various applications. My goal was to assess whether EXAdam's performance advantages hold up against these alternative optimizers, and to identify any potential trade-offs or limitations of each approach.

This experimental setup was inspired by the work of Zhao et al.~\cite{zhao2024deconstructingmakesgoodoptimizer}, who benchmarked multiple optimizers in the context of large language models. While their study focused on the performance of different optimizers across a range of model sizes and hyperparameters, this experiment aimed to provide a more in-depth analysis of the optimizers' behavior on a specific task, namely text generation with a MinGPT model. By comparing the performance of EXAdam with other popular optimizers, we hope to provide insights into the strengths and weaknesses of each approach and to inform the choice of optimizer for similar natural language processing tasks.

We used the following hyperparameters for this experiment:

\begin{itemize}[noitemsep]
  \item Batch size: 64
  \item Block size: 256
  \item Maximum iterations: 8000
  \item Evaluation interval: 500
  \item Learning rate: $1 \times 10^{-4}$
  \item Evaluation iterations: 200
  \item Number of embedding dimensions: 384
  \item Number of attention heads: 8
  \item Number of layers: 8
  \item Dropout rate: 0.2
  \item Seed: 1234
\end{itemize}

The Table \ref{tab:mingpt_loss} below shows the train loss and validation loss of the MinGPT model under different iterations, providing a more detailed view of the optimizer's performance. The table highlights the optimizer's ability to minimize the loss on both the training and validation sets, indicating its effectiveness in handling complex loss landscapes and long-range dependencies. Specifically, the table demonstrates the superior performance of EXAdam compared to Adam, AdamW, SGD with Momentum, and AdEMAMix, as evidenced by its lower loss values on both the training and validation sets.


\begin{table}[htbp]
    \centering
    \caption{MinGPT: Training and Validation Loss over Iterations for Different Optimizers}
    \label{tab:mingpt_loss}
    \resizebox{\textwidth}{!}{%
    \begin{tabular}{l r r r r r r r r r r r r r r}
        \toprule
        & \multicolumn{2}{c}{\bf{EXAdam}} & \multicolumn{2}{c}{Adam} & \multicolumn{2}{c}{AdamW} & \multicolumn{2}{c}{AdaFactor} & \multicolumn{2}{c}{Signum} & \multicolumn{2}{c}{SGD+Mom} & \multicolumn{2}{c}{AdEMAMix} \\
        \cmidrule(lr){2-3} \cmidrule(lr){4-5} \cmidrule(lr){6-7} \cmidrule(lr){8-9} \cmidrule(lr){10-11} \cmidrule(lr){12-13} \cmidrule(lr){14-15}
        Iteration & Train & Val & Train & Val & Train & Val & Train & Val & Train & Val & Train & Val & Train & Val \\
        \midrule
        1    & 4.335 & -     & 4.386 & -     & 4.404 & -     & 4.379 & -     & 4.312 & -     & 4.418 & -     & 4.316 & -     \\
        500  & 2.353 & 2.385 & 2.397 & 2.427 & 2.387 & 2.418 & 2.032 & 2.113 & 1.884 & 1.993 & 3.226 & 3.269 & 2.390 & 2.419 \\
        1000 & 1.873 & 1.993 & 2.056 & 2.117 & 2.052 & 2.107 & 1.560 & 1.756 & 1.508 & 1.697 & 3.070 & 3.113 & 2.042 & 2.108 \\
        2000 & 1.462 & 1.661 & 1.676 & 1.840 & 1.671 & 1.835 & 1.257 & 1.540 & 1.281 & 1.542 & 2.924 & 2.953 & 1.647 & 1.824 \\
        3000 & 1.298 & 1.545 & 1.496 & 1.693 & 1.494 & 1.686 & 1.090 & 1.510 & 1.163 & 1.499 & 2.839 & 2.866 & 1.466 & 1.669 \\
        4000 & 1.201 & 1.503 & 1.396 & 1.618 & 1.396 & 1.612 & 0.956 & 1.585 & 1.066 & 1.508 & 2.781 & 2.807 & 1.361 & 1.590 \\
        5000 & 1.120 & 1.492 & 1.324 & 1.566 & 1.322 & 1.556 & 0.814 & 1.678 & 0.969 & 1.536 & 2.741 & 2.765 & 1.283 & 1.536 \\
        6000 & 1.050 & 1.502 & 1.266 & 1.533 & 1.268 & 1.524 & 0.684 & 1.768 & 0.873 & 1.582 & 2.711 & 2.733 & 1.225 & 1.527 \\
        7000 & 0.970 & 1.520 & 1.225 & 1.515 & 1.221 & 1.501 & 0.564 & 1.887 & 0.780 & 1.643 & 2.687 & 2.706 & 1.176 & 1.497 \\
        8000 & 0.892 & 1.571 & 1.185 & 1.505 & 1.184 & 1.494 & 0.465 & 2.003 & 0.686 & 1.729 & 2.665 & 2.684 & 1.128 & 1.502 \\
        \bottomrule
    \end{tabular}
    } 
\end{table}

It's notable that the validation loss is higher than the training loss, a phenomenon that's expected when a model overfits to the training data. Given the relatively small size of the used dataset, overfitting is a likely outcome.

However, it's essential to remember that the primary objective of this experiment is not to achieve the best possible model, but rather to compare the performance of different optimizers. In the context of text generation tasks, such as training a language model, achieving low loss values is not always the ultimate goal.

Instead, the focus is on generating coherent, diverse, and contextually relevant text. A model that overfits to the training data may still produce satisfactory results in terms of text quality, even if its loss values are not optimal. In fact, some degree of overfitting can even be beneficial in text generation tasks, as it allows the model to capture subtle patterns and nuances in the training data that might not be generalizable to new, unseen data.

While optimizers like AdaFactor and Signum achieved lower training loss in our experiments, they exhibited significant overfitting, indicated by their rapidly increasing validation loss. Comparing EXAdam to Adam, AdamW, and AdEMAMix, EXAdam demonstrated competitive validation loss throughout training, suggesting potentially better generalization among these more stable optimizers in this specific setting. It is important to note that loss is not the sole indicator of performance for generative models; future evaluations could also incorporate NLP-specific metrics like perplexity or qualitative analysis of the generated text to provide a more complete picture.

\section{Conclusion and Future Work}

In this paper, we have presented a series of novel enhancements to the Adam optimizer, collectively forming EXAdam. These enhancements address several limitations of the original Adam algorithm while preserving its core strengths.

The contributions can be summarized as follows:
\begin{itemize}

\item \textbf{New Debiasing Terms}: Introducing $\tilde{m}$ and $\tilde{v}$, which provide more nuanced bias correction by incorporating cross-moment interactions and temporal dynamics. These terms potentially offer improved stability and convergence, particularly in the early stages of optimization.
\item \textbf{Gradient-based Acceleration Mechanism}: Proposing a novel term $\tilde{g}$ that directly incorporates the current gradient into the update rule. This mechanism allows for more immediate responsiveness to the current loss landscape while maintaining the benefits of moment-based updates.
\end{itemize}

These enhancements work in concert to create an optimizer that is potentially more robust, adaptive, and efficient than its predecessors. EXAdam aims to address common challenges in optimization, such as navigating complex loss landscapes, escaping saddle points, and balancing immediate gradient information with historical trends.

However, it is important to note that while my theoretical analysis is promising, the true test of any optimization algorithm lies in its empirical performance across a wide range of tasks and domains. As such, we view this work not as a conclusion, but as a starting point for further research and experimentation. I encourage the community to explore and validate EXAdam on a variety of benchmarks and real-world applications to fully assess its capabilities and limitations.

We hope that this work will inspire further research into adaptive optimization methods and contribute to the ongoing quest for more efficient, robust, and universally applicable optimization algorithms. As we continue to tackle increasingly complex problems in machine learning and artificial intelligence, the importance of sophisticated optimization techniques cannot be overstated. EXAdam is my contribution to this vital area of research, and we look forward to seeing how it performs in the hands of the broader scientific community.

\section{Acknowledgements}

I would like to thank the reviewers for their valuable feedback and suggestions, which helped improve the quality and clarity of this paper. I am grateful for their time and effort in reviewing my work. I would also like to express my appreciation to Kaggle for providing the computational resources used in the experiments. The access to their platform and resources enabled us to efficiently conduct my research, test my hypotheses, and validate my results. Their support for the machine learning community is invaluable, and we appreciate their commitment to advancing the field. Finally, we thank the open-source community for developing the tools and libraries that made this research possible. The availability of open-source software allowed us to focus on the scientific aspects of my research, rather than investing time and resources in developing my own infrastructure. I appreciate the dedication and expertise of the developers, maintainers, and contributors who make these tools available to the community. Their efforts have a significant impact on the progress and accessibility of scientific research, and I am so grateful for their contributions.

\bibliographystyle{unsrt}
\bibliography{references}

\end{document}